%% file: main.tex
\pdfoutput=1
\documentclass[11pt]{article}

\usepackage{emnlp2021}

\usepackage{times}
\usepackage{latexsym}

\usepackage{xspace}
\usepackage{graphicx}
\usepackage{amsmath}
\usepackage{arydshln}
\usepackage{multirow}

\usepackage{enumitem}
\usepackage{booktabs}
\usepackage{fontawesome}

\usepackage[T1]{fontenc}
\usepackage[utf8]{inputenc}

\usepackage{microtype}

\usepackage{color}
\definecolor{darkgreen}{RGB}{43,163,39}
\renewcommand\fbox{\fcolorbox{red}{white}}

\iftrue
    \newcommand{\ancr}[1]{\textcolor{cyan}{}}
    \newcommand{\pradeepcr}[1]{\textcolor{blue}{}}
    \newcommand{\mattcr}[1]{\textcolor{teal}{}}
\else
    \newcommand{\ancr}[1]{\textcolor{cyan}{\bf\small [AN: #1]}}
    \newcommand{\pradeepcr}[1]{\textcolor{blue}{\bf\small [PD: #1]}}
    \newcommand{\mattcr}[1]{{\textcolor{teal}{\bf \small [Matt: #1]}}}
\fi

\iftrue
    \newcommand{\an}[1]{\textcolor{cyan}{}}
    \newcommand{\pradeep}[1]{\textcolor{blue}{}}
    \newcommand{\matt}[1]{\textcolor{teal}{}}
\else
    \newcommand{\an}[1]{\textcolor{cyan}{\bf\small [AN: #1]}}
    \newcommand{\pradeep}[1]{\textcolor{blue}{\bf\small [PD: #1]}}
    \newcommand{\matt}[1]{{\textcolor{teal}{\bf \small [Matt: #1]}}}
\fi

\newcommand{\eg}{\textit{e.g., \xspace}}
\newcommand{\ie}{\textit{i.e., \xspace}}

\newcommand{\textbl}[1]{\textcolor{blue}{#1}} 
\newcommand\lt[1]{{\lstinline+#1+}} 

\title{Mitigating False-Negative Contexts in Multi-Document\\ Question Answering with Retrieval Marginalization}

\author{Ansong Ni$^\diamondsuit$\thanks{\enskip Majority of the work done as an intern at AI2.}\quad Matt Gardner$^\clubsuit$\quad Pradeep Dasigi$^{\clubsuit}$ \\ 
  $^\diamondsuit$Department of Computer Science, Yale University\\
  $^\clubsuit$Allen Institute for AI\\
  \texttt{ansong.ni@yale.edu}\\ \texttt{\{mattg,pradeepd\}@allenai.org}}

\begin{document}
\maketitle
\begin{abstract}
Question Answering (QA) tasks requiring information from multiple documents often rely on a retrieval model to identify relevant information for reasoning. The retrieval model is typically trained to maximize the likelihood of the labeled supporting evidence. However, when retrieving from large text corpora such as Wikipedia, the correct answer can often be obtained from multiple evidence candidates. Moreover, not all such candidates are labeled as positive during annotation, rendering the training signal weak and noisy. This problem is exacerbated when the questions are unanswerable or when the answers are Boolean, since the model cannot rely on lexical overlap to make a connection between the answer and supporting evidence. We develop a new parameterization of set-valued retrieval that handles unanswerable queries, and we show that marginalizing over this set during training allows a model to mitigate false negatives in supporting evidence annotations. We test our method on two multi-document QA datasets, IIRC and HotpotQA. On IIRC, we show that joint modeling with marginalization improves model performance by 5.5 F1 points and achieves a new state-of-the-art performance of 50.5 F1. We also show that retrieval marginalization results in 4.1 QA F1 improvement over a non-marginalized baseline on HotpotQA in the fullwiki setting.\footnote{Code available at \url{https://github.com/niansong1996/retrieval_marginalization}.}

\end{abstract}

\section{Introduction}

\begin{figure}[t!]
    \centering
    \includegraphics[width=\columnwidth]{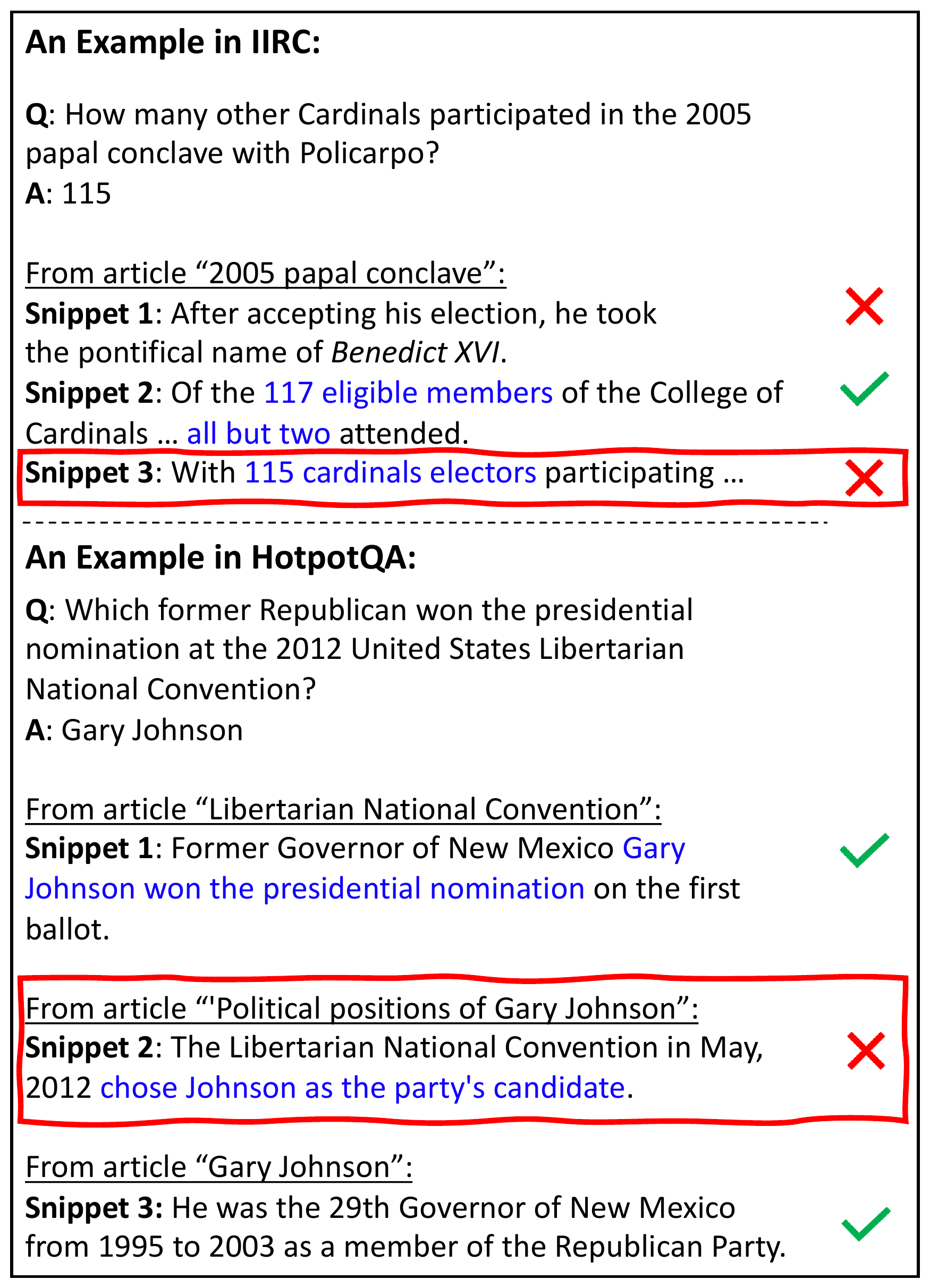}
    \caption{Examples of false-negative contexts in multi-document QA. Equivalent information is marked in \textcolor{blue}{blue}, and only the snippets with "\textcolor{darkgreen}{\faCheck}" are annotated as gold evidence. False negatives are \fbox{outlined in red} and both are retrieved by our proposed framework.
    % \mattcr{It's not obvious by looking at the x's and checkmarks that we're supposed to intuit a difference between snippet 1 and snippet 3 (one is a true negative, one is a false negative).  Changing something about the labeling here would be better. }
    % \ancr{fixed}
    }
    \label{fig:alternative-context}
\end{figure}

% \pradeep{I think the whole paper would be clearer if you consistently called the snippets that you are retrieving "evidence" or "evidence candidates", and the input to the QA model "context". }
Multi-document question answering refers to the task of answering questions that require reading multiple documents, extracting relevant facts, and reasoning over them. Systems built for this task typically involve retrieval and reasoning components that work in tandem. The retrieval component needs to extract information from the documents that is suitable for the reasoning model to perform the end-task effectively.
Recent advances in reading comprehension have resulted in models that have been shown to answer questions requiring complex reasoning types such as bridging, comparison \cite{Asai2020Learning, fang-etal-2020-hierarchical} or even arithmetic \cite{ran2019numnet, nmn:iclr20}, \emph{given adequate context}. However, when the context needs to be retrieved from a large text corpus (\eg Wikipedia), the performance of such reading comprehension models is greatly affected by the quality of the retrieval model.
Given supervision at all stages (\ie document, supporting evidence and answer), it is common to build retrieval and reasoning models independently and connect them as a pipeline at test time. In this case, the retrieval and reasoning models are usually trained to maximize the likelihood of labeled supporting evidence snippets and the answer given the gold context respectively.

However, in a multi-document QA setting, it is common to have some relevant snippets not marked as gold. Two such examples are shown in \autoref{fig:alternative-context}. In the first example, only snippet 2 is marked as gold evidence, and consequently snippets 1 and 3 are treated as negative examples during retrieval. This is problematic because unlike snippet 1 which is actually irrelevant, snippet 3 is not only useful, but provides an even more direct way to derive the correct answer since it does not require a subtraction. Similarly, in the second example two evidence snippets from different documents contain the same information, thus at least two contexts can be used to answer this question, yet only one of them is labeled as being a positive example for the training objective. 
We define these contexts that contain non-gold snippets and can still be used to answer the questions as \textit{alternative contexts}. Alternative contexts are inevitable when datasets are created from large corpora, because it is prohibitively expensive to exhaustively annotate all possible contexts.  These alternative contexts are false negatives during training and lead to a noisy and weak learning signal, even with this "fully-supervised" setup.

We design a training procedure for handling these false negatives, as well as cases where retrieval should fail (\ie when the question is unanswerable).
% Targeting this problem and to make use of the alternative contexts, we propose to jointly learn the retrieval model and reasoning model \matt{This whole sentence is awkward and vague; consider instead ``We propose to structure our model and training algorithm to account for these false negatives.''}. 
Specifically, we assign probabilities to documents, evidence candidates, and potential answers with parameterized models, and marginalize over a set of potential contexts by combining top retrieved evidence from each document, allowing the model to score false negatives highly. To make the marginalization feasible, we decompose the retrieval problem into document selection and evidence retrieval and show how we can still model contexts as sets.
% Adding to the maximum likelihood objective, the models are also optimized towards retrieving the alternative contexts and deriving correct answers from those non-gold contexts \matt{Not sure this sentence as worded really adds anything.  Perhaps instead: ``This marginalization gives the model freedom to assign high probability to false negatives, instead of forcing the model to assign them low probability, which would be a confusing learning signal.''}.
We evaluate our model on two multi-document QA datasets: IIRC~\citep{ferguson2020iirc} and HotpotQA~\citep{yang2018hotpotqa}. 
% Our method improves the final QA F1 performance by 4.8 and 8.9 points from the traditional pipeline approaches on IIRC and HotpotQA fullwiki setting, respectively. 
We see 2.8 and 4.8 F1 point improvement on IIRC and HotpotQA respectively by jointly modeling our proposed set-valued retrieval and the reasoning steps, and a further 2.7 and 4.1 F1 point improvement respectively by using retrieval marginalization.
Our final result of 50.5 F1 on the test set of IIRC represents a new state-of-the-art. 
% \pradeep{Our primary contribution is the marginalization. So, including improvements from joint modeling in the difference can be misleading. May be say something like ``'' Not sure if the differences I wrote in the last sentence are correct, especially for IIRC. Please check.} \an{okay, I fixed it.}
% Moreover, we achieve state-of-the-art results on IIRC with 50.6 answer F1, a 17.6 improvement compared to its baseline \matt{the 17.6 is somewhat disingenuous, as it's not a 17.6 point improvement relative to a comparable baseline. We care about science, not black-box numbers on leaderboards.  Instead: ``Our final result of 50.6 F1 on IIRC represents a new state-of-the-art on that dataset.''}.

\section{Multi-Document QA}
Here we formally describe the multi-document QA setting and highlight the two main challenges in this setting that our work attempts to address.
\paragraph{Problem Definition}\label{sec:problem-definition}
Multi-document question answering measures both the retrieval and reasoning abilities of a QA system. Given a question $q$ and a set of documents $\mathcal{D}=\{d^1,d^2,...,d^n\}$, each document containing a set of evidence snippets $d^i=\{s^i_1, s^i_2,...,s^i_{n_i}\}$, the goal of the model is to output the correct answer $a$. This task is typically modeled with a retrieval step, which locates a set of evidence $C=\{s^{i_1}_{j_1}, s^{i_2}_{j_2},...,s^{i_k}_{j_k}\}$ to formulate a context, and a reasoning step to derive the answer from such context $C$. Though such models can be learned with or without annotations on supporting evidence, we focus on the fully-supervised setting and assume supervision for all stages.
It is also common for such documents to have some internal structure (\eg hyperlinks in Wikipedia, citations for academic papers), which can be used to constrain the space of retrieval.

\paragraph{Inevitable False-negatives in Context Retrieval Annotations}\label{sec:challenges}

% \begin{table}
% \small
% \centering
% \begin{tabular}{lr}
% \toprule
% Total questions \# & 44 \\
% Total documents \# & 50 \\
% Documents w/ at least one alternative evidence & 27 \\
% Avg. \# alternative evidence per document\footnotemark & 1.14 \\
% Question w/ at least one alternative evidence & 28 \\
% \bottomrule
% \end{tabular}
% \caption{\textbf{Alternative evidence in IIRC.} Analysis is done on a small subset of the training set.}
% \label{tab:iirc-pilot}
% \end{table}
% \footnotetext{We count up to 5 alternative evidence snippets per document when measuring the average. Some documents contain many more alternative evidence snippets. An example would be when a question seeks information on the "winner of World War I" in relevant Wikipedia pages.}
Even when supporting evidence is annotated, we claim that the learning signal provided by those labels may be weak and noisy when retrieving from a large corpus 
such as Wikipedia. This is due to the redundancy of information in such large corpora: it is common to have multiple sets of evidence snippets that can answer the same question, as in \autoref{fig:alternative-context}. 
To quantify how often alternative contexts exist for the multi-document QA problem, we analyzed IIRC \cite{ferguson2020iirc}, an information seeking multi-document QA dataset. We sampled 50 answerable questions with their annotated gold context and manually checked if equivalent information can be found in sentences not labeled as supporting evidence, in the same document. We found that more than half of the questions have at least one alternative evidence, and on average \footnote{We count up to 5 alternative evidence snippets per document when measuring the average. Some documents contain many more alternative evidence snippets. An example would be when a question seeks information on the "winner of World War I" in relevant Wikipedia pages.}
there is more than one sentence we can find in the same document that contains the same information as the gold evidence. Note that it is also possible to have alternative evidence in a different document, which would further increase the frequency of questions with false-negative contexts. 
%\an{can remove the table and shrink this if we need more space.}\matt{I agree with removing the table. It looks funny to have a table with small numbers like that.  Just state the important numbers in the text.}

Due to the prevalence of such false-negative contexts, simply training the retrieval model to maximize the likelihood of the labeled supporting evidence would result in the models ignoring or even being confused by other unlabeled relevant information that could benefit the reasoning process. 
The problem is more severe when considering questions with Boolean answers or those that are unanswerable since the answers have no lexical overlap with corresponding evidence, making it harder to identify unlabeled yet relevant evidence snippets. 
Such false-negative context annotations are also inevitable in the data creation process, since the annotators will have to exhaustively search for evidence snippets from all relevant documents, which is rather impractical. As solving this problem is not typically feasible during data collection, we instead deal with it during learning.

\paragraph{Learning to Reason with Noisy Context}
Given retrieved supporting evidence as context, the second step of the problem is reading comprehension. Recently proposed models have shown promise in answering questions that require multi-hop \cite{Asai2020Learning, fang-etal-2020-hierarchical} or numerical \cite{ran2019numnet, nmn:iclr20} reasoning given small and sufficient snippets as contexts. However, the performance of such models degenerates rapidly when they are evaluated in a pipeline setup and attempt to reason with retrieved contexts that are potentially noisy and incomplete. For instance, \citet{ferguson2020iirc} found that the performance of reasoning models dropped 39.2 absolute F1 when trained on gold contexts and evaluated on retrieved contexts. 
This is mainly because the model is exposed to much less noise at training time than at test time. 
\paragraph{Handling Retrieval for Unanswerable Questions}
It is especially challenging when we consider unanswerable questions since it is possible to have seemingly relevant documents that are missing key information.\footnote{An example would be looking for the birth year of some person when such information is not presented even in the Wikipedia article titled with their name.} This is common for an information-seeking setting such as IIRC, where the question annotators are only given an initial paragraph to generate questions, with the actual content of linked documents being unseen.
Thus it raises the question of how to make use of such learning signal and correctly model the retrieval step for unanswerable questions.
% given initial paragraphs to generate questions, and they may label some links to other documents to be relevant but the answer annotators might override them after reading the linked documents by marking the question as unanswerable when the relevant information is not present. 

\section{Learning with Marginalization over Retrieval}
\begin{figure}
    \centering
    \includegraphics[width=1.0\linewidth]{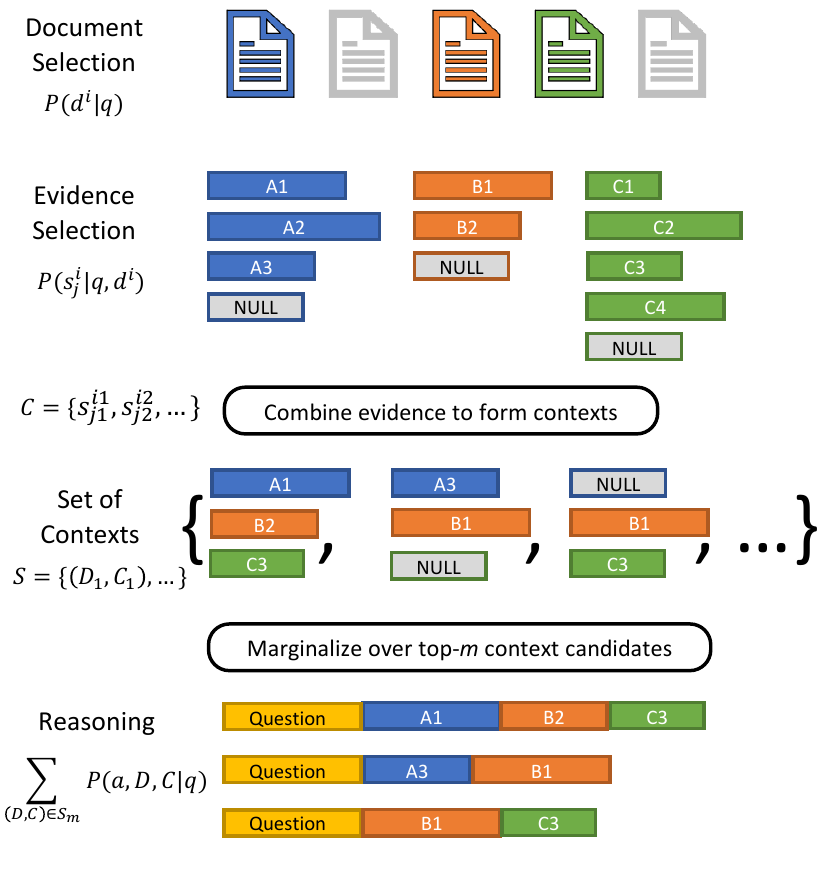}
    \caption{Our proposed framework of set-valued retrieval for handling context of unknown size.}
    % \matt{It might be a good idea to explicitly show the marginalization here.  You have it written in text, but it's not in the formulas.  E.g., replace $P(D,C,a|q)$ with $\sum_{D,C}P(D,C,a|q)$} \an{okay, does it look good now?}}
    \label{fig:framework}
\end{figure}
% \matt{You should talk about set-valued retrieval here first.  Going straight to this breakdown undersells your contribution.  That is, you need to say that fundamentally what we want is a model that selects a set of contexts C, and then marginalizes over that set.  But the contexts come from separate documents, and we don't know how many documents we should be looking at or how many sentences we need from each document.  We need an efficient way of parameterizing this set-valued random variable over contexts such that we can reasonable get a set for approximate marginalization.  To do this, we decompose the problem into document selection and evidence retrieval in a way that still lets us model sets.  This needs to be described up front, at least intuitively, before you go into the details.  }
% \matt{Some allusion to this should probably also show up in the intro, as it's one of the main contributions of this work.}
To address these challenges, we decompose the retrieval problem into document selection and evidence retrieval, leaving the handling of unanswerable questions entirely to the evidence retrieval component.  These two components together produce a probability distribution over sets of retrieved contexts, which we marginalize over during training to account for false negatives. Our general framework is illustrated in \autoref{fig:framework}.
\subsection{Modeling Multi-Document Retrieval}\label{sec:joint-retrieval}
As described in \autoref{sec:problem-definition}, the end product of retrieval for multi-document QA should be a set of evidence snippets from different documents $\{s^{i_1}_{j_1}, s^{i_2}_{j_2},...,s^{i_k}_{j_k}\}$ which are combined to be the reasoning context $C$. When a question is not answerable, $C$ is empty, and supervision of the model is not entirely straightforward.  The decision of where to look for evidence is separate from whether the necessary information is present, and a naive supervision that nothing should be retrieved risks erroneously telling the model that the place it chose to look was incorrect, possibly leading to spurious correlations between question artifacts and answerability judgments.  For this reason, we separate document selection from evidence retrieval, and we leave answerability determinations entirely to the evidence retrieval step.

\paragraph{Document selection} Given the question $q$ and a set of documents $\mathcal{D}=\{d^1, d^2,...,d^n\}$, to evaluate the relevance of each document $d^i$ and the question $q$, we first jointly encode them with a transformer-based model. To model the selection of documents as a set variable, a Sigmoid function ($\sigma$) is used to compute the document probability: 
%\an{maybe too much detail, can cut if we need more space} \pradeep{I agree. You can just define $x_{d^i}$ and $P(d^i | q)$ in text and remove the equations} \an{commented out for now, come back if we need more space}
\begin{flalign*}
    x_{d^i} &= \text{Encode}(d^i, q) \\
    P(d^i|q) &= \sigma(w_d^{\top} \cdot x_{d^i} + b_d)
\end{flalign*}
For simplification, we assume the selection of each document is independent, thus the joint probability of selecting a set of documents $D =\{d^{i_1}, d^{i_2},...,d^{i_k}\}, D\subseteq \mathcal{D}$ can be computed with:
% \ancr{I kept using $i$ because it was consistent throughout the notations and $j$ is already used in something else. But I also added a square bracket to make it clearer. Did the same thing for $\mathcal{G}_D$ under equation (4)}
\begin{flalign}\label{eq:doc-set-prob}
    P(D|q) &= \prod_{d^{i}\in D} P(d^{i}|q) \cdot \prod_{d^{i'} \in \mathcal{D} - D} (1-P(d^{i'}|q)) 
\end{flalign}
%Here function $T$ represents a joint encoding of the question and the document with pretrained contextual encoders.
% BERT \cite{devlin2019bert, liu2019roberta}\pradeep{just say contextual encoder instead of BERT}. 
%We discuss different settings of $T$ section \autoref{sec:exp_setup}.

\paragraph{Evidence retrieval} Given the set of selected documents $D$, the goal for the evidence retrieval model is to select the evidence snippets $s_j^i \in d^i$ that are relevant to question $q$ for each document $d^i \in D$. To model the relevance between an evidence snippet and the question, we first use pretrained language models to obtain a joint embedding of the concatenated question-evidence input. \an{also too much detail}
% \matt{remove this from here, as it's repeated below}:
% \begin{flalign*}
%     x_{s_j^i} &= \text{Encode}([s_j|| q])
% \end{flalign*}
To simplify the problem while approximating the set of evidence snippets as context, we only take one evidence snippet from each document.
% \pradeep{I changed some ``evidences'' to ``evidence snippets''. Please make sure it's consistent in the paper.} \pradeep{also ``one evidence snippet'' instead of ``one evidence''}  
In addition, we allow the evidence retrieval model to retrieve nothing from a document by predicting \texttt{NULL}, a special token which is artificially added to the end of every document. 
This is essential for our modeling since it places the responsibility of determining sentence-level relevance solely on the evidence retrieval step and allows it to reject the proposal from the document model by selecting the \texttt{NULL} option, which is useful especially for unanswerable questions.
% This is essential for our modeling since it gives the evidence retrieval model freedom to refute the decision from document-level model, which we found to be crucial especially in an information-seeking setting.\an{added this (<--) sentence, not sure if it makes sense} \pradeep{Since we mention this idea earlier, you could just say ``It is by adding these \texttt{NULL} tokens that we make the evidence retrieval step responsible for unanswerability prediction instead of the document selection step.''} 
Finally, we model the probability of an evidence snippet being retrieved given its document as: 
%\pradeep{You used comma to represent concatenation in document selection, and double pipes here. Use the same notation consistently.}
\begin{flalign*}
    x_{s_j^i} &= \text{Encode}([s_j|| q]) \\
    P(s_j^i|q, d^i) &= \underset{s_j^i\in d^i}{\operatorname{Softmax}}(w_s^{\top} \cdot x_{s_j^i} + b_s) 
    %P(s_j^i|q, d^i) &= \frac{exp(w_s^{\top} \cdot x_{s_j^i} + b_s)}{\sum_{s_k^i \in d_i} exp(w_s^{\top} \cdot x_{s_k^i} + b_s)}
\end{flalign*}
Here we can derive the joint probability of a set of evidence snippets $C=\{s^{i_1}_{j_1}, s^{i_2}_{j_2},...,s^{i_k}_{j_k}\}$ being retrieved as context:
\begin{flalign}\label{eq:joint-retrieval-prob}
    P(D,C|q) &= P(D|q) \cdot P(C|D,q) \\
    % &= \prod_{d^{i}\in D^{-}} (1-P(d^{i}|q)) \\
    % &\cdot \prod_{d^{i}\in D}\prod_{s_j^i\in d^i} P(d^{i}|q) \cdot P(s_j^i | d^i, q)
    &= P(D|q) \prod_{s_j^i\in C, s_j^i\in d^i, d^{i}\in D} P(s_j^i | d^i, q)\nonumber
\end{flalign}

\subsection{Joint Modeling with Marginalization} \label{sec:joint-modeling-with-marginalization}
With the retrieved context $C$, the final step is to predict the answer. For this part, we use existing reading comprehension models that take a question and relatively small context and output a probability distribution of its answer predictions. 
The retrieved sentences in the context are simply concatenated and treated as context for the reading comprehension model (RC). Given the context $C$ and question $q$, the probability of the answer is defined as:
\begin{flalign}
    P(a|q, D, C) = \text{RC}(q, [s_{j_1}^{i_1}||s_{j_2}^{i_2}||...s_{j_k}^{i_k}])
\end{flalign}
% \pradeep{You did not define D yet.}
Now we can derive the joint probability of the answer and the retrieved context:
\begin{flalign*}
    &P(a,D,C|q) \\
    = &P(a|D,C,q)\cdot P(D,C|q) \\
    % = &\prod_{d^{i}\in D^{-}} (1-P(d^{i}|q)) \\
    % &\cdot \prod_{d^{i}\in D}\prod_{s_j^i\in d^i} P(a|D,C) \cdot P(d^i|q) \cdot P(s_j^i | d^i, q) 
    = &P(D|q) \prod_{s_j^i\in C, s_j^i\in d^i, d^{i}\in D} P(a|D,C) \cdot P(s_j^i | d^i, q) 
\end{flalign*}
% \matt{update with (1-P)}
% \matt{make sure the ordering of variables aligns with the formulas in your figure} \an{done}
With the objective to maximize the likelihood of the training set with supervision on gold $\bar{C}$, $\bar{D}$ and $\bar{a}$, the loss function is as in \autoref{eq:mle_obj}:
%\pradeep{Should be $\min_{\theta} - (\mathcal{G}_D + ...)$, because loss is something that should be minimized, right?  Or just call this the objective instead of ``loss''. If you choose to call it loss, Eq. 5, 6, and 7 need to use $\min_{\theta}$ too.} \an{changed all \max to \min}
\begin{flalign}\label{eq:mle_obj}
    &\min_\theta \mathcal{G}_{D} + \mathcal{G}_{C} + \mathcal{G}_{a}
\end{flalign}
where
\begin{flalign*}
    \mathcal{G}_{D} = &-\sum_{d^{i}\in \bar{D}} \log P(d^{i}|q) \\
    &- \sum_{d^{i'}\in \mathcal{D}-\bar{D}} \log(1-P({d^{i'}}|q))\\
    \mathcal{G}_{C} = &-\sum_{d^i\in \bar{D}}\sum_{s^i_j\in d^i, s^i_j\in \bar{C}}\log P(s^i_j|d^i,q) \\
    \mathcal{G}_{a} = &-\log P(\bar{a}|\bar{D},\bar{C},q)
\end{flalign*}
% \matt{update with (1-P)}

\paragraph{Marginalization over Retrieved Evidence}
As mentioned in 
% \pradeep{You don't need "section" if you have this squiggly thing here->} 
\autoref{sec:challenges}, the learning signals for $\{\mathcal{G}_{D}, \mathcal{G}_{C}, \mathcal{G}_{a}\}$ may be noisy and weak because the objectives in \autoref{eq:mle_obj} assume that given a question-answer pair $(q,a)$ there is only one set of gold context $\bar{C}$ that can derive the correct answer.
% \mattcr{There are some minor notational inconsistencies throughout the paper---you only sometimes use a bar, and you sometime use superscripts for documents and sometimes subscripts (see the Document Selection section).  Double check all of your notation and make sure it's consistent.} 
% \ancr{checked and fixed}
To augment the learning signal, we propose to add the weakly-supervised objective with marginalization over a set of alternative context $S=\{(D'_1,C'_1), (D'_2,C'_2),...,(D'_m,C'_m)\}$ given the selected documents $D$:
\begin{flalign}\label{eq:wsp_obj}
    \mathcal{G}_{M} &= -\log P(\bar{a}|q) \\ 
    &= -\log \sum_{(D'_i,C'_i)\in S}P(\bar{a},D'_i,C'_i|q) \nonumber
\end{flalign}
Ideally, we want the marginalization set $S$ to be all possible combinations of sentences in different documents, but this is infeasible for large text corpora. So here, we approximate the marginalization set by: 1) using only the top-ranked document set $D$, and 2) selecting only the top-$m$ contexts from each $d^i$ in $D$. 
% \matt{by (1) using only the top-ranked document set $D$, and (2) selecting only the top $m$ contexts from each $d^i$ in $D$. (and cut the rest of this sentence)} to be the top-$m$ retrieved context ranked by their joint probability output from the document selection and evidence retrieval model as in \autoref{eq:joint-retrieval-prob}.
% \pradeep{No need for the equations below. The text is clear enough.}\an{since we want to emphasize the "set-valued retrieval", maybe we should have an equation to define how those are generated?but I will cut this out if we are short on space}:
% \begin{flalign*}
%     (C', D') &= \{s^{i_1}_{j_1},...,s^{i_k}_{j_k}\} \\
%     &= \arg\max_{C, D} P(s^{i_k}_{j_k}|d^{i_k}, q)\cdot P(d^{i_k} | q)
% \end{flalign*}

However, not all of the contexts in the top-$m$ set $S$ are good alternative contexts, especially when the retrieval model is under-trained and performs poorly. We use a set of answer-type-dependent heuristics to determine whether a context $C$ is \textit{valid}:
% \mattcr{I think it makes more sense to have those heuristics inline here, so that it's more obvious that the next section makes sense.}
% \ancr{done}
(1) \textit{Span}: when the context has at least one span that matches the gold answer string;
(2) \textit{Number}: when the answer can be derived from the numbers in the context with the arithmetic operation supported by our RC model, or it is a span in the context;
(3) \textit{Yes/No/Unanswerable}: all contexts are considered valid.
Using these heuristics, we can divide the top $m$ retrieved context $S$ into two subsets $S_1, S_2 \subseteq S$, where $S_1$ contains all \textit{valid} contexts while the contexts in $S_2$ are invalid.

\paragraph{Auxiliary Loss for Invalid Context}
Because contexts in $S_2$ are not valid alternative contexts for obtaining the correct answer, we do not marginalize over contexts in this set. We can still use them during training, however, by formulating an auxiliary loss that encourages the RC model to predict the question as unanswerable (\ie $a=a_N$) given these invalid contexts: 
% \pradeep{$a_N$ is not defined. May be write it as $a=\texttt{NONE}$ instead?} \an{defined now}
\begin{flalign}\label{eq:ic_obj}
    \mathcal{G}_{a_N} = -\sum_{(D^*,C^*)\in S_2} \log P(a_N|D^*,C^*,q)
\end{flalign}
Note that here we do not use joint probability $P(a_N, D^*, C^*|q)$ since doing so would also encourage the retrieval models to retrieve irrelevant context for answerable questions. In this way, this auxiliary loss can also be viewed as augmenting the dataset with extra unanswerable question-context pairs for the RC model.

Our final learning objective of the joint modeling is the result of combining the objectives in \autoref{eq:mle_obj} , \autoref{eq:wsp_obj} and \autoref{eq:ic_obj}:
\begin{flalign}
    \min_\theta \mathcal{G}_{D}+\mathcal{G}_{C}+\mathcal{G}_a + \mathcal{G}_{M} + \alpha\cdot\mathcal{G}_{a_N}
\end{flalign}
% \pradeepcr{You may want to remove the parentheses in the equation above.}\ancr{I thought the parentheses emphasizes our contribution, but yeah, in math it doesn't make sense}
% \matt{Recommended: $\mathcal{G}_D+\mathcal{G}_C+\mathcal{G}_a+\mathcal{G}_M+\alpha\mathcal{G}_{a_N}$}\an{sounds good}
The only weight we tune in the objective is $\alpha$ to regulate the contribution of the loss from the invalid contexts the RC model encounters at training time. 
% \matt{Type mismatch here, as the loss is defined in terms of a weighting, but the text talks about number of questions.  That's not what the equation says.  They should match.}\an{I changed to ``distribution of loss'', does it work?}\pradeep{Reworded it further.}

\section{Experiments}\label{sec:exp_setup}
\input{tab-iirc-main-results}

\subsection{Datasets and Settings}
We test our method on two multi-document question answering datasets: IIRC and HotpotQA.

\paragraph{IIRC} \cite{ferguson2020iirc} is a dataset consisting of 13K information-seeking questions generated by crowdworkers who had access only to single Wikipedia paragraphs and the list of hyperlinks to other Wikipedia articles, but not the articles themselves. Given an initial paragraph, a model needs to retrieve missing information from multiple linked documents to answer the question. Since the question annotators can only see partial context, the questions and contexts containing the answers have less lexical overlap. The questions in IIRC may have one of four types of answers: 1) \textit{span}; 2) \textit{number} (resulting from discrete operations); 3) \textit{yes/no}; 4) \textit{none} (when the questions are unanswerable).

\paragraph{HotpotQA fullwiki} \cite{yang2018hotpotqa} consists of 113K questions and the contexts for answering those questions are a pair of Wikipedia paragraphs. In the \textit{fullwiki} setting, the model has access to 5.2M paragraph candidates and needs to retrieve relevant information from this corpus. We believe this open domain QA setting would provide a different perspective for studying false-negative contexts, especially across different documents. 
%\an{as we did find something in later analysis}

\subsection{Model Details}
Transformer-based pretrained language models are used to encode questions and contexts for retrieval and reasoning in our experiments. To be as comparable to previous methods as possible, we use \texttt{RoBERTa-base} for IIRC and \texttt{BERT-large-wwm} for HotpotQA fullwiki. For document selection in IIRC, we only rely on the initial paragraph with hyperlinks and not the content in linked documents themselves. On HotpotQA fullwiki, we follow \citet{Asai2020Learning} and bootstrap our document selection model from their recurrent graph retriever. 
For the reasoning part, we use NumNet+ \cite{ran2019numnet} for IIRC and \texttt{BERT-wwm} \cite{devlin2019bert} for HotpotQA. 
\ancr{added the following sentence}
Note that our general modeling framework is agnostic to the choices of specific models for retrieval and reasoning. We choose these models to experiment with since they are easy to use and present strong results as shown in previous work \citep{groeneveld-etal-2020-simple}.
More implementation details can be found in \autoref{sec:model-details}.

% \subsection{Identifying Valid Context}\label{sec:valid-context-heuristics}
% \ancr{Moved this part up from the appendix}
% As described in \autoref{sec:joint-modeling-with-marginalization}, only \textit{valid} contexts are added to the marginalization set, while invalid ones are used to compute the auxiliary loss. Depending on the answer types, we use different heuristics to identify the validity of a context for answering the question: \\
% \noindent \textbf{Span}: when the context has at least one span that matches the gold answer string; \\
% \noindent \textbf{Number}: when the answer can be derived from the numbers in the context with the arithmetic operation supported by our RC model, or it is a span in the context; \\
% \noindent \textbf{Yes/No}: all contexts are considered valid; \\
% \noindent \textbf{None}: all contexts are considered valid.
% \an{reformulated the following from \textbackslash itemize to plain text} For \textit{span} types, we judge by whether the context has at least one span that matches the gold answer string. As for \textit{numbers} as answers, we check if the answer can be derived from the numbers in the context with the arithmetic operations supported by our RC model, or if it is a span in the context. Lastly, with \textit{binary} and \textit{unanswerable} questions, all contexts are considered valid.

% \matt{this itemize takes up a \emph{ton} of space; you probably want to just keep this as a paragraph.  easy to describe this with a sentence or two, as you can combine descriptions.}:

\subsection{Evaluation Metrics}
For HotpotQA, we follow previous work \cite{yang2018hotpotqa, Asai2020Learning, xiong2020answering} and use F1 score and Exact Match (EM) for the answer (QA) and supporting facts (SP) prediction. Similarly, we report QA F1 and EM for IIRC as in \citet{ferguson2020iirc}. In addition, we define the following metrics for understanding the retrieval performance:
\textit{(1) Document selection F1 (Doc-F1)} measures the performance of the document retrieval model given the documents marked as gold; 
\textit{(2) Overall retrieval recall(Rt-Recall)} measures the retrieval ability of the overall retrieval system given the annotated set of evidence snippets.
Over the metrics above, our main goal is to improve question answering performance, which is best measured by QA F1.
% \matt{You probably want to say which metric you care about the most.  Something like, ``Our main goal is to improve question answering performance, which is best measured by QA F1.''  I'd probably start this section with that sentence, then define QA F1.  Then say, ``We additionally report other metrics that give insight into models' performance, including:''}

\subsection{Training Settings}
Since documents can contain up to hundreds of sentences, for efficient training of our evidence retrieval model, we downsample the negative examples to 7 for IIRC and 3 for HotpotQA. But no downsampling is done during inference.
For IIRC, we take $m=4$ for the top-$m$ context marginalization and take $m=5$ for HotpotQA. For the weight for invalid context loss, we use $0.5$ for IIRC\footnote{We performed a simple binary search and found 0.5 to work better than 0 or 1.} and $0$ for HotpotQA since it does not have unanswerable questions. 
% To be more memory and storage efficient, our joint model performs parameter sharing by using the same pretrained language models across different components. 
For memory and storage efficiency, we tie the pretrained language model weights among all the components in our joint model.
% \pradeepcr{Suggestion to rephrase as ``For memory and storage efficiency, we tie the pretrained language model weights among all the components in our joint model.''}\ancr{great suggestion}
%We evaluate and discuss the effect of this for a fair comparison with the pipeline models which cannot perform such parameter sharing in \autoref{sec:exp-analysis}.
The models are trained for 30 epochs for IIRC and 5 epochs for HotpotQA fullwiki. 
Our most expensive experiment takes about 1.5 days to run on two RTX 8000 (48GB) GPUs or one A100 (40GB) GPU, while a typical experiment takes about half of that computing power.\footnote{Note that our models are jointly trained, thus this is the total training cost for both retrieval and reasoning.}

\section{Main Results on IIRC}
% \subsection{Overall Effectiveness}\label{sec:overall-effectiveness}\matt{Remove this subsection and just have it be underneath section 5.}
\autoref{tab:iirc-main-result} shows our main results on IIRC. We can see that our proposed joint model with marginalization outperforms the pipeline model by 5.2 and 4.8 points for QA exact match and F1 score, respectively. While the 17.6 point improvement over the baseline system seems large, the correct point of comparison for our contribution in this work is our pipeline system, which is simply an improved version of the pipeline used by the baseline system.\footnote{The main differences include using RoBERTa instead of BERT during retrieval, adding a \texttt{NULL} option for evidence retrieval and having a binary classifier for binary questions.}
Another comparison worth noticing is that despite the large improvement on the QA side, the retrieval performance is slightly lower than its pipeline counterpart. Our hypothesis is that our trained joint model with marginalization can better utilize the alternative contexts that are not marked as gold and derive the correct answers based on them. Since the retrieval performance is compared with only the annotated evidence, the gain from alternative contexts cannot be reflected in these numbers. 
We explore this hypothesis in \autoref{sec:exp-analysis} and some examples are shown in \autoref{sec:qual-analysis} and \autoref{sec:qual-analysis-2}.

To further understand the effectiveness of joint modeling with marginalization comparing to the pipeline model, we breakdown the QA performance by different answer types in \autoref{tab:iirc-breakdown}.
% \mattcr{I've been removing these autorefs because they result in "Tab. 2" instead of "Table 2", which is unsightly.  Also they give a space between the \S and the number, which is not how \S is typically used.  I'd suggest removing them throughout, or finding a way to customize the behavior of autoref.} \ancr{I've customized the autoref behavior, but the words "Table", "Section" are able hyperlinked, I think it should be okay?}
Our proposed method yields large performance gains on unanswerable questions, and those with binary and numerical answers. As we discussed in \autoref{sec:challenges}, since the retrieval model cannot rely on lexical overlap between contexts and correct answers for these types of questions, it is harder for it to learn from the false negatives, and the reasoning model trained in a pipeline is more susceptible to noise in the retrieved context. 
We also notice that the QA F1 on span-type questions drops 1.2 points; we think this is because the auxiliary loss we have on invalid contexts slightly altered the distribution to favor the unanswerable questions. To confirm this, we removed the auxiliary loss and its QA F1 on span questions went back up to 48.7 points.

\input{tab-iirc-type-breakdown}

\subsection{Analysis}\label{sec:exp-analysis}
\paragraph{Effectiveness of retrieval marginalization}
\autoref{tab:iirc-ablation} shows that training with marginalization improves the final QA F1 performance by 2.7 points, while doing slightly worse in terms of retrieving annotated context.
To go beyond pure numbers and explain why modeling with retrieval marginalization results in better final QA performance, 
we analyzed 50 questions where the model with marginalization correctly answered a question that the model without marginalization missed, and 50 questions where the opposite was true.  We found that in 24\% of the cases where the marginalization model was correct, it relied on non-gold evidence to make its prediction, while this was only true 4\% of the time for the model without marginalization.
This suggests that marginalization over retrieval improves the QA performance by retrieving alternative contexts that can help reasoning. We show some specific examples of this in \autoref{sec:qual-analysis} and \autoref{sec:qual-analysis-2}.

\iffalse
\begin{table}
\small
\centering
\begin{tabular}{lrr}
\toprule
\textbf{Reason} & \textbf{marg.} & \textbf{no marg.} \\
\midrule
Document selection & 2\% & 6\% \\
Labeled evidence retrieval & 28\% & 40\% \\
Alternative evidence retrieval & 24 \% & 4\% \\
Spurious context & 6\% & 4\% \\
Reading comprehension & 50\% & 56\% \\
\bottomrule
\end{tabular}
\caption{\textbf{Cross-model advantage analysis on retrieval marginalization}. ``Spurious context'' refers to a model deriving correct answer from retrieved context without any logical connection. Categories are \textit{not} mutually exclusive.}
\label{tab:iirc-marg-study1}
\end{table}
\fi

\input{tab-iirc-ablation}

\paragraph{Effectiveness of invalid context loss}
\footnotetext{It is also a must because marginalization depends on joint modeling and the auxiliary loss depends on having a marginalization set.}
Without the auxiliary loss on the subset of invalid contexts during marginalization, we observe a 1.4 points decrease in the QA performance. On further inspection, we found that the main reason was the performance on the unanswerable questions, which decreases 8.9 points in F1 (not shown in the table).
%This shows the effectiveness of this part of the objective in identifying unanswerable questions.

\paragraph{Effectiveness of joint modeling}
We also explore the setting where both marginalization and joint modeling are taken away from our model. This is similar to the pipeline setting but we minimize the sum of fully supervised losses
from all three models and the pretrained language model weights are shared. 
The difference between rows 3 and 4 in \autoref{tab:iirc-ablation} illustrates the performance improvements from joint modeling alone, which is 2.8 QA F1.
We believe this is largely due to the fact that when the model is jointly trained, the reasoning model is dynamically adapting to the noisy retrieval results, which makes it more resilient to noise at test time. 
%\an{We are missing a point here: joint modeling can also mitigate false-negative problem, it's just marginalization does the job better, maybe add something about this?}.

\section{Results on HotpotQA}
\label{sec:hotpot-results}

\input{tab-hotpot-results}

The general effectiveness of retrieval marginalization and joint modeling on HotpotQA fullwiki is displayed in \autoref{tab:hotpot-results}. We observe a 4.8 QA F1 improvement with joint modeling and a further 4.1 points improvement with retrieval marginalization. Similar to IIRC, joint modeling and retrieval marginalization improve final QA performance despite inferior retrieval scores evaluated against annotated supporting evidence. To better understand how our proposed methods alter the retrieval and reasoning steps, we compare the output from our full model as well as the version without marginalization and joint modeling. Interestingly, while we found most alternative sources of evidence are from the same document in IIRC, the alternative evidence our full model locates for HotpotQA fullwiki is mostly from different documents. We believe this is because HotpotQA fullwiki considers far more documents (\ie 5.2M in the whole Wikipedia) than IIRC (\ie less than 20 linked documents) for each question, but HotpotQA typically use the introductory paragraph instead of the whole document, giving less chance for inner-document alternatives. Concrete examples are shown in \autoref{sec:qual-analysis} and \autoref{sec:qual-analysis-2}

While the purpose of these experiments is to show the effectiveness of our proposed methods for mitigating the false negatives in HotpotQA rather than competing with state-of-the-art models, we do list some results from previous work in \autoref{tab:hotpot-results} to help put ours in context. 
With joint modeling and retrieval, our full model achieves a QA F1 of 71.2 and EM of 58.6, surpassing several strong baselines presented by previous work, though there is still a non-trivial gap with state-of-the-art models. Potential improvements that are orthogonal to our contributions include modeling document selection as a sequence and condition selection based on previous selected documents as done by \citet{Asai2020Learning} and \citet{xiong2020answering}, and using stronger language models (\eg ELECTRA \cite{clark2019electra}). Note that all three modules of our framework are simple classification models on top of BERT, and that the formulation of our proposed set-valued retrieval and marginalization are model-agnostic. 
% \an{not sure how to properly send the message that "it can be apply to previous methods to improve their performance" without being attacked by "why don't you do this and try to get SOTA already".}\pradeep{I added a phrase to convey this.}

% We also note that our supporting fact prediction (SP) performance is much lower than previous work. In addition to the effect of alternative evidence mentioned above, another reason is that most previous frameworks use full paragraphs for reasoning and use predicted answer as an input for selecting supporting evidence. This is feasible because typically only the first paragraph of each Wikipedia document is considered for HotpotQA fullwiki. When we consider actual documents, evidence selection needs to precede reasoning step since a concatenation of multiple documents is too long and noisy for common reasoning models \cite{clark2017simple}.
% \an{maybe cite some document QA paper} \matt{This whole paragraph has run-on sentences that I'm not sure how to fix, because it's not entirely clear to me what they are trying to say.} \an{not sure if it looks good now.} \an{removing this paragraph after a discussion with Pradeep.}

\section{Qualitative Analysis}
\label{sec:qual-analysis}
\input{tab-case-study-main}
\ancr{Moved this part up from the appendix, along with Tab. 5, added some more analysis}
As mentioned in \autoref{sec:exp-analysis} and \autoref{sec:hotpot-results}, the reason for our model to achieve higher QA scores though having lower scores in matching annotated evidence is that our model locates false-negative evidence that can be used as alternatives for reasoning. In \autoref{tab:case-study-main}, we show two of such alternatives our model found in the development set for IIRC and HotpotQA fullwiki. From the IIRC example, we can see that our model is able to find alternative evidence in different sentences of the same gold document (\ie \textit{"Alexander Hamilton"}) or even in a non-gold document (\ie \textit{"Aaron Burr"}), mitigating false-negatives annotations in sentence or document retrieval. The HotpotQA example shows that in multi-hop reasoning, key evidence is not exclusively in documents titled with bridging entity (\ie \textit{"E Street band"}), but also sometimes included in a document for related third entity (\ie \textit{"David Sancious"}) as well. This indicates that such false-negative contexts can be more prevalent in multi-hop QA settings.

\section{Related Work}
% \matt{Perhaps have a note here somewhere that this confirms prior analysis in the NQ and TyDiQA papers that says humans typically have low recall in this setting, though they were focused on dealing with issues in evaluation.  In this work, we're focused on dealing with issues in training motivated by the same problem.  Also perhaps reasonable to put this off to a related work section.}
\paragraph{False-negative contexts in QA} 
In terms of dealing with false negatives in retrieved texts for question answering, the most similar prior work to ours is by~\citet{clark2017simple}. However, they focused only on span-type answers while we apply similar methods to more complex reasoning types. 
False negative contexts are not exclusive to IIRC or HotpotQA, but are a rather common issue when scaling up QA to document or multi-document level. In prior analysis of TyDiQA \cite{clark-etal-2020-tydi} and Natural Questions \cite{kwiatkowski-etal-2019-natural}, it is suggested that humans typically have low recall on finding supporting evidence, though they focused on dealing with such issues in evaluation while we focus on altering the training process. Such issue was also found by previous work on HotpotQA, as \citet{xiong2020answering} noted that many of the "errors" in their document retrieval model are actually valid alternative contexts. The false-negative contexts can also lead to false-negative annotations of answer spans. In some previous work \cite{chen2017reading, hu-etal-2019-retrieve, Asai2020Learning},\matt{third ref is broken} it was shown effective to manually add distantly-supervised examples in the reasoning model's training, while we try to solve this problem from the retrieval part.

\paragraph{Joint models for retrieval and reasoning} 
More recent work such as ORQA \cite{lee2019latent} and Dense Passage Retrieval \cite{karpukhin2020dense} focus on modeling retrieval for question answering. But their main focus is to develop efficient neural retrieval systems that help to scale up QA to a large corpus, which is orthogonal to our contribution in mitigating false-negative contexts.
% Work from~\citet{xiong2020answering} focuses on retrieval for multi-hop question answering, they note in their error analysis that a large portion of the retrieval errors were actually alternative contexts.\matt{You already said this above.}
Some recent works also investigate the possibility of modeling retrieval as a latent task and enabling end-to-end training. REALM \cite{guu2020realm} used a neural retriever to augment language models with external knowledge for question answering. While REALM also uses a maximum marginal likelihood objective, it only marginalizes over retrieved documents. However, our model marginalizes over a set of contexts consisting of evidence snippets from different documents to account for set-valued retrieval of variable sizes, which is crucial for multi-document QA.
RAG \cite{lewis2020retrieval} adopted a similar method as REALM, but for sequence generation tasks. RAG can be adapted to perform multi-hop question answering \cite{xiong2020answering}, and we directly compared our results with multi-hop RAG on HotpotQA fullwiki in the experiments.
Finally, our work leverages marginalization over latent variables to deal with weak and noisy supervision signals, which is reminiscent of using maximum marginal likelihood for training weakly supervised semantic parsers~\citep[][among others]{berant2013semantic,krishnamurthy2017neural}.

\section{Conclusion}
We proposed a new probabilistic model for retrieving set-valued contexts for multi-document QA and show that training the QA model with marginalization over this set can help mitigate the false negatives in evidence annotations. Experiments on IIRC and HotpotQA fullwiki show that our proposed framework can learn to retrieve unlabeled alternative contexts and improves QA F1 by 5.5 on IIRC and 8.9 on HotpotQA.
% \matt{isn't the improvement on IIRC bigger than this?  you can use the full improvement over the pipeline model here.} \an{okay, done}

\section*{Acknowledgements}
\ancr{added the acks}The authors would like to thank Tushar Khot, Hannaneh Hajishirzi, James Ferguson, Dragomir Radev, and the anonymous reviewers for the useful discussion and comments. 

\section*{Ethical Considerations}
We address the ethical considerations of this work from the following perspectives:
\paragraph{Misuse potential} Our models are built specifically for answering factoid questions on Wikipedia. We think it is unlikely for our method to be misused for other domains.
\paragraph{Failed modes} In our case, failing to answer a user-issued question may result in incorrect or misleading information. Thus we should be careful when put our systems into practical use.
\paragraph{Computation power} Our most expensive experiment takes about 1.5 days to run on two RTX 8000 (48GB) GPUs or one A100 (40GB) GPU, while a typical experiment takes about half of that computing power. While conducing experiments, we made effort to take advantage of technologies such as mixed-precision training to shorten our training time under the same experiment setting and save power consumption.

% Entries for the entire Anthology, followed by custom entries
\bibliography{anthology,custom}
\bibliographystyle{acl_natbib}

\clearpage
\appendix

\section{Model Implementation Details}
\label{sec:model-details}
Since the two datasets where we conduct our experiments are different in terms of document length and structure, reasoning types from questions and possible answer types, here we describe the dataset-specific implementation details for IIRC and HotpotQA. 

\subsection{Document Selection}
% \pradeep{Merge 4.5 with this section and have a subsection each describing how the setup varies between the two datasets.}\an{4.5 is more about hyperparameter settings and describes things like downsampling and parameter sharing which are not really dataset-specific setting}
Though IIRC and both settings from HotpotQA can be seen as multi-document QA problems, the documents are structured differently. Accordingly we handle the document selection part differently for the two datasets, namely using different $\text{Encode}(\cdot)$ functions mentioned in section \autoref{sec:joint-retrieval}. But note that the outputs are all a set of documents (or paragraphs for HotpotQA) with their probabilities, so other parts of the framework remain the same.

\paragraph{Link prediction for IIRC.} For IIRC, an initial paragraph $p$ is given and we need to follow certain links in it, so the document selection problem can be translated into a link prediction problem given $p$ and $q$. So for IIRC, we define $\text{Encode}(\cdot)$ as the BERT embedding of the concatenated question-paragraph sequence at the position of the link $l_i$ to the document $d_i$:
\begin{flalign*}
    \text{Encode}_{\text{IIRC}}(d^i,q) = \text{BERT}_{[l_i]} ([q||p])
\end{flalign*}
For IIRC, since we do not know how many links needs to be followed to answer the questions, we use $P(d^i|q,p)=0.5$ as a threshold for document selection. \\

\paragraph{HotpotQA fullwiki.} To select paragraphs for HotpotQA, we simply concatenate the question and the first 64 tokens of the candidate paragraph $d^i$, run it through BERT and take the embedding of the separation token:
\begin{flalign*}
    \text{Encode}_{\text{Hotpot}}(d^i,q) = \text{BERT}_{[\text{SEP}]} ([q||d^i])
\end{flalign*}
We further follow \citet{Asai2020Learning} and apply their trained recurrent retriever to retrieve a small subset of relevant paragraphs $\mathcal{D}'$ with the highest score. We choose to use this model because they are the best performing model on HotpotQA fullwiki setting with public code. For a better learning signal, we manually add the paragraphs marked as gold if they are not already included in $\mathcal{D}'$ but we only use $\mathcal{D}'$ at test time for a fair comparison under the fullwiki setting. 

\subsection{Reading Comprehension Models}
We use existing reading comprehension models for both IIRC and HotpotQA. NumNet+ \cite{ran2019numnet} is used for IIRC since it can handle numerical reasoning. We augment the model by adding binary and unanswerable as two additional question types to its question type classification model and further introduce a binary classification model for outputting "Yes" and "No" when a question is classified as binary type.
For HotpotQA, to handle questions with binary answers, we append "yes or no" to the start of the retrieved context to transform its reasoning part to a pure span-prediction problem. Then
we follow \citet{devlin2019bert} and append two linear layers to the contextualized embeddings from transformer-based language models, and they are used to separately model the starting and ending position of the span. 

\section{Retrieved Alternative Contexts}
\label{sec:qual-analysis-2}
\input{tab-case-study-iirc}
\input{tab-case-study-hotpot}
Here we show more examples of alternative contexts retrieved by our proposed methods, for the development set for IIRC in \autoref{tab:case-study-iirc} and for HotpotQA fullwiki in \autoref{tab:case-study-hotpot}.

\end{document}

%% file: tab-iirc-main-results.tex
\begin{table*}
\centering
\small
\begin{tabular}{lccccccccc}
\toprule
&\multicolumn{4}{c}{\textbf{Dev}}& &\multicolumn{4}{c}{\textbf{Test}}\\\cline{2-5}\cline{7-10}
&\multicolumn{2}{c}{\textbf{Retrieval}} &\multicolumn{2}{c}{\textbf{Reasoning}}&
&\multicolumn{2}{c}{\textbf{Retrieval}} &\multicolumn{2}{c}{\textbf{Reasoning}}\\\cline{2-5}\cline{7-10}
\textbf{Model} & Doc-F1 & Rt-Recall & QA EM & QA F1 && Doc-F1 & Rt-Recall & QA EM & QA F1 \\
\midrule
Baseline (NumNet+) & - & - & 29.6 & 33.0 & & - & - & 27.7 & 31.1 \\ \hdashline
Pipeline (NumNet+) &  & \ & 41.7 & 45.8 & & & & 41.3 & 44.3\\
$^\dagger$Pipeline (T5-Large) & \textbf{88.7} & \textbf{62.8} & 44.2 & 47.4 & & 90.8 & \textbf{67.9} & 37.8 & 41.0\\
$^\dagger$Pipeline (PReasM-Large) & & & 46.5 & 50.0 & & & & 42.0 & 45.1\\
\midrule
Ours (NumNet+) & 87.3 & 62.0 & \textbf{46.9} & \textbf{50.6} & & \textbf{91.0} & 67.5 & \textbf{47.4} & \textbf{50.5} \\
\bottomrule
\end{tabular}
\caption{\textbf{Main results on IIRC.} \ancr{changed a lot for this table} "Baseline" refers to the performance reported in \citet{ferguson2020iirc} and "-" denotes that no results are available. Work marked with $^\dagger$ is by \citet{yoran2021turning}, which appeared after our initial submission. All pipeline models shares the same retrieval model and its output thus the same retrieval performance.
% Models below the dash line share the same architecture for retrieval and reasoning models and only differs in learning objectives.
}
\label{tab:iirc-main-result}
\end{table*}

%% file: tab-iirc-type-breakdown.tex
\begin{table}
\small
\centering
\begin{tabular}{lcl}
\toprule
\textbf{Answer Types (\%)} & \textbf{Pipeline} & \textbf{Joint Model} \\
\midrule
None (26.7\%)     & 49.6     & 62.2 (+12.6)        \\
Yes/No (9.8\%)    & 52.3     & 62.5 (+10.2)        \\
Span (45.6\%)     & 48.4     & 47.2 (-1.2)        \\
Number (17.9\%)   & 30.0     & 35.6 (+5.6)        \\
\midrule
All (100\%)       & 45.8     & 50.6 (+4.8)       \\
\bottomrule
\end{tabular}
\caption{\textbf{IIRC answer F1 breakdown by answer types.} Number in the brackets refer to the absolute improvement from the pipeline model.}
\label{tab:iirc-breakdown}
\end{table}

%% file: tab-iirc-ablation.tex
\begin{table}
\small
\centering
\begin{tabular}{l|cccc}
\toprule
\textbf{Model} & \textbf{Doc-F1} & \textbf{Rt-Recall} & \textbf{QA F1}\\
\midrule
full model         & 87.3   & 62.0      & \textbf{50.6}  \\\hdashline
 - invalid context loss &  87.5   & \textbf{62.5} & 49.2  \\
\;\;\;- marginalization      & \textbf{87.8}   & 62.2      & 47.9  \\
\;\;\;\; - joint modeling       & \textbf{87.8}   & 62.4      & 45.1 \\
\bottomrule
\end{tabular}
\caption{\textbf{Marginalization and other ablations on IIRC.} Note that the removal of parts (noted by "-") from the full model is \textit{accumulative}\footnotemark. The last setting is equivalent to a pipeline model with shared RoBERTa weights.}
\label{tab:iirc-ablation}
\end{table}

%% file: tab-hotpot-results.tex
\begin{table}
\small
\centering
\begin{tabular}{lcccc}
\toprule
&\multicolumn{2}{c}{\textbf{SP}} &\multicolumn{2}{c}{\textbf{QA}}\\
\textbf{Model} & EM & F1 & EM & F1\\
\midrule
full model             & 14.3   & 60.7  & \textbf{58.6} & \textbf{71.2} \\\hdashline
- marginalization      & 18.2   & 63.1  & 54.4 & 67.1 \\
\;\;- joint modeling   & \textbf{28.0}   & \textbf{66.0}  & 50.2 & 62.3  \\
\midrule
\textbf{Previous work}\\
SR-MRS \cite{nie2019revealing} & 39.9 & 71.5 & 46.5 & 58.8  \\
T-XH \cite{zhao2019transformer} & 42.2 & 71.6 & 50.2 & 62.4 \\
HGN \cite{fang-etal-2020-hierarchical} & 50.0 & 76.4 & 56.7 & 69.2 \\
\citet{Asai2020Learning} & 49.2 & 76.1 & 60.5 & 73.3 \\
MDR \cite{xiong2020answering} & \textbf{57.5} & \textbf{80.9} & \textbf{62.3} & \textbf{75.3}  \\
RAG\footnotemark \cite{lewis2020retrieval} & - & - & 51.2 & 63.9 \\
\bottomrule
\end{tabular}
\caption{\textbf{HotpotQA answer F1 on the development set.} Best performance in our ablation and from previous work are in \textbf{bold}. "-" denotes that no results are available.}
\label{tab:hotpot-results}
\end{table}
\footnotetext{Here we compare with Multi-hop RAG, an adaptation of RAG \cite{lewis2020retrieval} to HotpotQA fullwiki by \citet{xiong2020answering}.} 

%% file: tab-case-study-main.tex
\begin{table*}[t]
    \centering
    \footnotesize
    \begin{tabular}{l}
    \toprule
        \textsc{IIRC Example} \\
    \midrule
        \textbf{Question}: Who won the famous duel that took place in Weehawken? \\
        \textbf{Answer}: Aaron Burr \\
        \textbf{Evidence in initial paragraph}: \\
        \underline{Hudson County, New Jersey}: Weehawken became notorious for duels, including the nation's most famous between \\Alexander Hamilton and Aaron Burr in 1804. \\
        \textbf{Gold evidence}: \\
        \underline{Alexander Hamilton}: Burr took careful aim and shot first, and Hamilton fired while falling, after \textcolor{blue}{being struck by Burr's bullet}. \\
        \textbf{Predicted alternative evidence}: \\
        1. \underline{Alexander Hamilton}: Taking offense, Burr challenged him to a duel on July 11, 1804, in which \textbl{Burr shot and mortally} \\ \textbl{wounded Hamilton}, who died the following day. \\
        2. \underline{Aaron Burr}: Both men fired, and \textbl{Hamilton was mortally wounded} by a shot just above the hip. \\
    \midrule
        \textsc{HotpotQA fullwiki Example} \\
    \midrule
        \textbf{Question}: What band did both Delores Holmes and an inductee to the Rock and Roll Hall of Fame sing in?  \\
        \textbf{Answer}: E Street Band \\
        \textbf{Gold evidence}: \\
        \underline{Garry Tallent} : Tallent was inducted as \textcolor{blue}{a member of the E Street Band into the Rock and Roll Hall of Fame}. \\
        \textbf{Predicted alternative evidence}: \\
        1. \underline{David Sancious} : In 2014, Sancious was inducted into \textcolor{blue}{the Rock and Roll Hall of Fame as} \textcolor{blue}{a member of the E Street Band}.  \\
        2. \underline{E Street Band} : The band was inducted into \textcolor{blue}{the Rock and Roll Hall of Fame} in 2014. \\
        \textbf{Other evidence in common}: \\
        \underline{Delores Holmes} : She was best known for her years as backup singer for the Bruce Springsteen Band during 1969 to 1972,\\ the last grouping before the E Street Band. \\
    \bottomrule
    \end{tabular}
    \caption{Examples of alternative evidence retrieved for IIRC (up) and HotpotQA fullwiki (down) by our model. Document titles are \underline{underlined} and key information is marked in \textcolor{blue}{blue}.}
    \label{tab:case-study-main}
\end{table*}

%% file: tab-case-study-iirc.tex
\begin{table*}[t]
    \centering
    \footnotesize
    \begin{tabular}{l}
    \toprule
    %     \textbf{Question}: Who won the famous duel that took place in Weehawken? \\
    %     \textbf{Answer}: Aaron Burr \\
    %     \textbf{Evidence in initial paragraph}: \\
    %     \underline{Hudson County, New Jersey}: Weehawken became notorious for duels, including the nation's most famous between \\Alexander Hamilton and Aaron Burr in 1804. \\
    %     \textbf{Gold evidence}: \\
    %     \underline{Alexander Hamilton}: Burr took careful aim and shot first, and Hamilton fired while falling, after \textcolor{blue}{being struck by Burr's bullet}. \\
    %     \textbf{Predicted alternative evidence}: \\
    %     1. \underline{Alexander Hamilton}: Taking offense, Burr challenged him to a duel on July 11, 1804, in which \textbl{Burr shot and mortally} \\ \textbl{wounded Hamilton}, who died the following day. \\
    %     2. \underline{Aaron Burr}: Both men fired, and \textbl{Hamilton was mortally wounded} by a shot just above the hip. \\
    % \midrule
        \textbf{Question}: In what year was the football club joined by Haycock for one season in 1910 first founded? \\
        \textbf{Answer}: 1884 \\
        \textbf{Evidence in initial paragraph}: \\
        \underline{Fred Haycock}: After a couple of years he ... and then Southern League clubs Luton Town and Portsmouth, before returning \\to the Football League with Lincoln City in 1910. \\
        \textbf{Gold evidence}: \\
        \underline{Lincoln City F.C.}: After the disbanding of Lincoln Rovers (formerly Lincoln Recreation) \textbl{in 1884}, \textbl{Lincoln City FC was formed}. \\
        \textbf{Predicted alternative evidence}:  \\
        \underline{Lincoln City F.C.}: Previously, Lincoln City had played at the nearby John O'Gaunts ground since \textbl{the club's 1884 inception}. \\
    \midrule
        \textbf{Question}: Where was the representative who gave approval to read excerpts from McCaughey's op-ed born? \\
        \textbf{Answer}: Waterloo, Iowa \\
        \textbf{Evidence in initial paragraph}: \\
        \underline{Death panel}: On July 27, excerpts from the McCaughey's op-ed were read, with approval, by Representative (Rep.) Michele \\ Bachmann (R-MN) on the floor of the U.S. House of Representatives. \\
        \textbf{Gold evidence}: \\
        \underline{Michele Bachmann}: Bachmann was born Michele Marie Amble \textbl{in Waterloo, Iowa}. \\
        \textbf{Predicted alternative evidence}:  \\
        \underline{Michele Bachmann}: Michele Marie Amble was born in \textbl{Waterloo, Iowa} on April 6, 1956, to Norwegian-American parents David\\ John Amble (1929–2003) and "Arlene" Jean Amble (née Johnson) (born c. 1932). \\
    \midrule
        \textbf{Question}: Did the singer Nicholls opened for at the De La Warr Pavilion release any studio albums? \\
        \textbf{Answer}: Yes \\
        \textbf{Evidence in initial paragraph}: \\
        \underline{Danni Nicholls}: Nicholls also opened as support for Lucinda Williams at the De La Warr Pavilion Bexhill-on-Sea for one of only \\ two shows in the UK and completed ... \\
        \textbf{Gold evidence}: \\
        \underline{Lucinda Williams}: In 1988, she released \textbl{her self-titled album, Lucinda Williams}. \\
        \textbf{Predicted alternative evidence}:  \\
        \underline{Lucinda Williams}: The American folk/rock band Augustana references the musician in the song "Meet You There," on their \\ \textbl{studio album Can't Love, Can't Hurt}. \\
    \bottomrule
    \end{tabular}
    \caption{Examples of alternative evidence retrieved for IIRC by our model. Document titles are \underline{underlined} and key information is marked in \textcolor{blue}{blue}.}
    \label{tab:case-study-iirc}
\end{table*}

%% file: tab-case-study-hotpot.tex
\begin{table*}[t]
    \centering
    \footnotesize
    \begin{tabular}{l}
    \toprule
        \textbf{Question}: What race track in the midwest hosts a 500 mile race every May? \\
        \textbf{Answer}: Indianapolis Motor Speedway \\
        \textbf{Gold evidence}: \\
        \underline{1957 Indianapolis 500} : The 41st International 500-Mile Sweepstakes was \textcolor{blue}{held at the Indianapolis Motor Speedway} \\ on Thursday, May 30, 1957. \\
        \textbf{Predicted alternative evidence}:  \\
        1. \underline{1974 Indianapolis 500} : The 58th 500 Mile International Sweepstakes was held \textcolor{blue}{at the Indianapolis Motor Speedway} in \\Speedway, Indiana on Sunday, May 26, 1974. \\
        2. \underline{1977 Indianapolis 500} : The 61st International 500 Mile Sweepstakes was held \textcolor{blue}{at the Indianapolis Motor Speedway} in \\Speedway, Indiana on Sunday, May 29, 1977. \\
        \textbf{Other evidence in common}: \\
        \underline{Indianapolis Motor Speedway} : The Indianapolis Motor Speedway is an automobile racing circuit located in Speedway, \\ Indiana, (an enclave suburb of Indianapolis) in the United States. \\
    \midrule
        \textbf{Question}: What is the middle name of the actress who plays Bobbi Bacha in Suburban Madness? \\
        \textbf{Answer}: Ann \\
        \textbf{Gold evidence}: \\
        \underline{Bobbi Bacha} : Bobbi Bacha is a Texas Private Investigator portrayed in 2004 TV Sony Pictures Movie "Suburban Madness" \\\textcolor{blue}{played by actress Sela Ward.} \\
        \textbf{Predicted alternative evidence}:  \\
        \underline{Suburban Madness} : Suburban Madness is an American crime drama television film, based on a true story, \textcolor{blue}{starring Sela} \\ \textcolor{blue}{Ward as PI Bobbi Bacha} of Blue Moon Investigations. \\
        \textbf{Other evidence in common}: \\
        \underline{Sela Ward} : Sela Ann Ward (born July 11, 1956) is an American actress, author and producer, best known for her roles on \\ television beginning in the early '80s. \\
    \bottomrule
    \end{tabular}
    \caption{Examples of alternative evidence retrieved for HotpotQA fullwiki by our model. Document titles are \underline{underlined} and key information is marked in \textcolor{blue}{blue}.}
    \label{tab:case-study-hotpot}
\end{table*}